\pdfoutput=1
\documentclass[10pt]{article}

\usepackage[letterpaper, left=1in, right=1in, 
top=1in, bottom=1in]{geometry}
\usepackage{breakcites}
\usepackage[dvipsnames]{xcolor}
\usepackage[hidelinks]{hyperref}
\hypersetup{
	colorlinks=true,
	linkcolor=blue!70!black,
	citecolor=blue!70!black,
	urlcolor=blue!70!black}

\usepackage{microtype,bm}
\usepackage{xspace}
\usepackage{algorithm}
\usepackage{verbatim}
\usepackage[noend]{algpseudocode}
\usepackage{mathtools}
\usepackage{amsmath}
\usepackage{bbm}
\usepackage{amsfonts}
\usepackage{amssymb}
\usepackage{xpatch}
\usepackage{enumitem}
\usepackage{amsthm}

\usepackage[nointegrals]{wasysym}

\theoremstyle{definition}  

\newtheorem{lemma}{Lemma}

\theoremstyle{plain}

\xpatchcmd{\proof}{\itshape}{\normalfont\proofnameformat}{}{}
\newcommand{\proofnameformat}{\bfseries}

\usepackage{prettyref}
\newcommand{\pref}[1]{\prettyref{#1}}

\newcommand{\savehyperref}[2]{\texorpdfstring{\hyperref[#1]{#2}}{#2}}
\newrefformat{eq}{\savehyperref{#1}{\textup{(\ref*{#1})}}}
\newrefformat{eqn}{\savehyperref{#1}{Equation~\ref*{#1}}}
\newrefformat{lem}{\savehyperref{#1}{Lemma~\ref*{#1}}}
\newrefformat{obs}{\savehyperref{#1}{Observation~\ref*{#1}}}
\newrefformat{def}{\savehyperref{#1}{Definition~\ref*{#1}}}
\newrefformat{line}{\savehyperref{#1}{line~\ref*{#1}}}
\newrefformat{thm}{\savehyperref{#1}{Theorem~\ref*{#1}}}
\newrefformat{corr}{\savehyperref{#1}{Corollary~\ref*{#1}}}
\newrefformat{sec}{\savehyperref{#1}{Section~\ref*{#1}}}
\newrefformat{app}{\savehyperref{#1}{Appendix~\ref*{#1}}}
\newrefformat{ass}{\savehyperref{#1}{Assumption~\ref*{#1}}}
\newrefformat{ex}{\savehyperref{#1}{Example~\ref*{#1}}}
\newrefformat{fig}{\savehyperref{#1}{Figure~\ref*{#1}}}
\newrefformat{alg}{\savehyperref{#1}{Algorithm~\ref*{#1}}}
\newrefformat{rem}{\savehyperref{#1}{Remark~\ref*{#1}}}
\newrefformat{conj}{\savehyperref{#1}{Conjecture~\ref*{#1}}}
\newrefformat{prop}{\savehyperref{#1}{Proposition~\ref*{#1}}}
\newrefformat{proto}{\savehyperref{#1}{Protocol~\ref*{#1}}}
\newrefformat{prob}{\savehyperref{#1}{Problem~\ref*{#1}}}
\newrefformat{claim}{\savehyperref{#1}{Claim~\ref*{#1}}}

\def\ddefloop#1{\ifx\ddefloop#1\else\ddef{#1}\expandafter\ddefloop\fi}
\def\ddef#1{\expandafter\def\csname 
	bb#1\endcsname{\ensuremath{\mathbb{#1}}}}
\ddefloop ABCDEFGHIJKLMNOPQRSTUVWXYZ\ddefloop
\def\ddef#1{\expandafter\def\csname 
	#1\endcsname{\ensuremath{{\bm{#1}}}}}
\ddefloop ABCDEFGHIJKLMNOPQRSTUVWXYZ\ddefloop
\def\ddef#1{\expandafter\def\csname 
	#1\endcsname{\ensuremath{{\bm{#1}}}}}
\ddefloop abcdefghijklmnopqrstuvwxyz\ddefloop
\def\ddefloop#1{\ifx\ddefloop#1\else\ddef{#1}\expandafter\ddefloop\fi}
\def\ddef#1{\expandafter\def\csname 
	b#1\endcsname{\ensuremath{\mathbf{#1}}}}
\ddefloop ABCDEFGHIJKLMNOPQRSTUVWXYZ\ddefloop
\def\ddef#1{\expandafter\def\csname 
	c#1\endcsname{\ensuremath{\mathcal{#1}}}}
\ddefloop ABCDEFGHIJKLMNOPQRSTUVWXYZ\ddefloop
\def\ddef#1{\expandafter\def\csname 
	f#1\endcsname{\ensuremath{\mathfrak{#1}}}}
\ddefloop ABCDEFGHIJKLMNOPQRSTUVWXYZ\ddefloop
\def\ddef#1{\expandafter\def\csname 
	h#1\endcsname{\ensuremath{\widehat{#1}}}}
\ddefloop ABCDEFGHIJKLMNOPQRSTUVWXYZ\ddefloop
\def\ddef#1{\expandafter\def\csname 
	hc#1\endcsname{\ensuremath{\widehat{\mathcal{#1}}}}}
\ddefloop ABCDEFGHIJKLMNOPQRSTUVWXYZ\ddefloop
\def\ddef#1{\expandafter\def\csname 
	t#1\endcsname{\ensuremath{\widetilde{#1}}}}
\ddefloop ABCDEFGHIJKLMNOPQRSTUVWXYZ\ddefloop
\def\ddef#1{\expandafter\def\csname 
	tc#1\endcsname{\ensuremath{\widetilde{\mathcal{#1}}}}}
\ddefloop ABCDEFGHIJKLMNOPQRSTUVWXYZ\ddefloop

\newcommand{\EE}{\mathbb{E}}
\newcommand{\RR}{\mathbb{R}}

\newcommand{\bzeros}{{\bm{0}}}
\newcommand{\bones}{\bm{1}}

\newcommand{\inner}[1]{\langle #1 \rangle}

\renewenvironment{proof}{\par\noindent{\bf Proof\ }}{\hfill\BlackBox\\[2mm]}
\newcommand{\BlackBox}{\rule{1.5ex}{1.5ex}}

\newcommand{\todoc}[1]{\ignorespaces}

\newcommand{\Diag}[1]{\text{Diag}\prn*{#1}}

\newcommand{\mymat}[1]{\prn*{\begin{array}{cccccccccccccccccc} #1 
\end{array}}}

\DeclarePairedDelimiter{\brk}{[}{]}
\DeclarePairedDelimiter{\crl}{\{}{\}}
\DeclarePairedDelimiter{\prn}{(}{)}
\DeclarePairedDelimiter{\nrm}{\|}{\|}

\usepackage{nicefrac}

\newsavebox\CBox 

\usepackage{color}
\usepackage{soul}

\newcommand{\grad}[1]{{\nabla {#1}}}
\newcommand{\hes}[1]{{\nabla^2\!{#1}}}
\newcommand{\is}[1]{{[#1]}}

\newcommand{\dd}{\mbox{$\;|\;$}}

\newcommand{\real}{{\mathbb{R}}}

\newcommand{\intg}{{\mathbb{Z}}}

\newcommand{\mc}[1]{{\mathbb{#1}}}

\newcommand{\On}[1]{{\text{\rm O}({#1})}}
\newcommand{\ON}{\On{d}}
\newcommand{\GLn}[1]{{\text{\rm GL}({#1})}}
\newcommand{\GLN}{\GLn{d}}
\newcommand{\Hn}[1]{{\text{\rm H}_{#1}}}
\newcommand{\HN}{\Hn{d}}
\newcommand{\defoo}{~\dot{=}~}

\newcommand{\sgn}{\mbox{sgn}}
\newcommand{\arr}{{\rightarrow}}

\newcommand{\NN}{\mathbb{N}}

\newtheorem{prop}[lemma]{Proposition}

\theoremstyle{definition}

\newtheorem{exams}[lemma]{Examples}

\theoremstyle{remark}

\usepackage{tikz}					
\usepackage{ifthen}

\tikzset{%
	box/.style={
		align=center, minimum size=1cm, inner 
		sep=0pt, font=\tiny\bfseries,
		draw=black, #1
	},
	c1/.style={fill=yellow},
	c2/.style={fill=blue},
	c3/.style={fill=green},
	c3/.style={fill=red},
}

\newcommand{\blockmatrix}[9]{
	\draw[draw=#4,fill=#5] (0,0) rectangle( #1,#2);
	\ifthenelse{\equal{#6}{true}}
	{
		\draw[draw=#7,fill=#8] (0,#2) -- (#9,#2) -- ( #1,#9) -- ( #1,0) -- ( #1 
		- #9,0) -- (0,#2 -#9) -- cycle;
	}
	{}
	\draw ( #1/2, #2/2) node { #3};
}

\newcommand{\mblockmatrix}[4][none]{
	\begin{tikzpicture} 
	\ifthenelse{\equal{#1}{none}}
	{
		\blockmatrix{#2}{#3}{#4}{none}{none}{false}{none}{none}{0.0}
	}
	{
		\definecolor{fillcolor}{rgb}{#1}
		\blockmatrix{#2}{#3}{#4}{none}{fillcolor}{false}{none}{none}{0.0}
	}
	\end{tikzpicture}
}

\newcommand{\fblockmatrix}[4][none]{
	\begin{tikzpicture} 
	\ifthenelse{\equal{#1}{none}}
	{
		\blockmatrix{#2}{#3}{#4}{black}{none}{false}{none}{none}{0.0}
	}
	{
		\definecolor{fillcolor}{rgb}{#1}
		\blockmatrix{#2}{#3}{#4}{black}{fillcolor}{false}{none}{none}{0.0}
	}
	\end{tikzpicture}
}

\newcommand{\dblockmatrixSD}[6][none]{
	\begin{tikzpicture} 
	\draw[draw=black,fill=none] (0,0) rectangle( #2,#3);
	\definecolor{fillcolor}{rgb}{#1};
	\draw[draw=black,fill=fillcolor] (0,#3) -- (#5,#3) -- ( #2,#5) -- ( #2,0) 
	-- ( #2 - #5,0) -- (0,#3 -#5) -- cycle;
	\draw ( #2/2, #3/2) node { #4};
	\draw ( #2*0.25, #3*0.25) node {#6};
	\draw ( #2*0.75, #3*0.75) node { #6};
	\end{tikzpicture}
}

\newcommand{\dblockmatrixSDMO}[7][none]{
	\begin{tikzpicture} 
	\draw[draw=black,fill=none] (0,0) rectangle( #2,#3);
	
	\definecolor{upperleft}{rgb}{#1};
	\draw[draw=black,fill=upperleft] (0,#3) -- (#5,#3) -- ( #2- #7,#5+#7) -- ( 
	#2 -  #7,#7) -- ( #2 - #5 - #7 ,#7) -- (0,#3 -#5) -- cycle;
	
	\definecolor{lowerleft}{rgb}{1,1,0.8};
	\draw[draw=black,fill=lowerleft] (0,0) rectangle (#2-#7,#7);
	
	\definecolor{upperright}{rgb}{0.8,1,0.8};
	\draw[draw=black,fill=upperright] (#2-#7,#7) rectangle (#2,#3);
	
	\definecolor{lowerright}{rgb}{0.8,0.8,1};
	\draw[draw=black,fill=lowerright] (#2-#7,0) rectangle (#2,#7);
	\draw ( #2*0.5-#7*0.5, #3*0.5+#7*0.5) node { #4};
	\draw ( #2*0.25-#7*0.25, #3*0.25 + #7*0.75) node {#6};
	\draw ( #2*0.75 - #7*0.75, #3*0.75 + #7*0.25) node { #6};
	\draw ( #2*0.5 - #7*0.5, #7*0.5) node {$\gamma$};
	\draw ( #2 - #7*0.5, #3*0.5+ #7*0.5) node {$\delta$};
	\draw ( #2 - #7*0.5, #7*0.5) node {$\epsilon$};
	\end{tikzpicture}
}

\newcommand{\dblockmatrixSDMOOO}[7][none]{
	\begin{tikzpicture} 
	\draw[draw=black,fill=none] (0,0) rectangle( #2,#3);
	
	\definecolor{upperleft}{rgb}{#1};
	\draw[draw=black,fill=upperleft] (0,#3) -- (#5,#3) -- ( #2- #7,#5+#7) -- ( 
	#2 -  #7,#7) -- ( #2 - #5 - #7 ,#7) -- (0,#3 -#5) -- cycle;
	
	\definecolor{lowerleft}{rgb}{1,1,0.8};
	\draw[draw=black,fill=lowerleft] (0,0) rectangle (#2-#7,#7);
	
	\definecolor{upperright}{rgb}{0.8,1,0.8};
	\draw[draw=black,fill=upperright] (#2-#7,#7) rectangle (#2,#3);
	
	\definecolor{lowerright}{rgb}{0.8,0.8,1};
	\draw[draw=black,fill=lowerright] (#2-#7,0) rectangle (#2,#7);
	\draw ( #2*0.5-#7*0.5, #3*0.5+#7*0.5) node { #4};
	\draw ( #2*0.25-#7*0.25, #3*0.25 + #7*0.75) node {#6};
	\draw ( #2*0.75 - #7*0.75, #3*0.75 + #7*0.25) node { #6};
	\draw ( #2*0.5 - #7*0.5, #7*0.5) node {$\gamma$};
	\draw ( #2 - #7*0.5, #3*0.5+ #7*0.5) node {$\delta$};
	\definecolor{lowerdia}{rgb}{0.8,0.8,0.8};
	\draw[draw=black,fill=lowerdia] (#2-#7,#7) -- (#2-#7+#5,#7) -- ( #2,#5) 
	-- ( 
	#2 ,0) -- ( #2 - #5  ,0) -- (#2-#7,#7 -#5) -- cycle;
	\draw ( #2 - #7*0.5, #7*0.5) node {$\epsilon$};
	\draw ( #2-#7*0.25 , #7*0.75) node {$\zeta$};
	\draw ( #2 - #7*0.75 , #7*0.25) node { $\zeta$};
	\end{tikzpicture}
	
}

\newcommand{\diagonalblockmatrix}[5][none]{
	\begin{tikzpicture} 
	
	\ifthenelse{\equal{#1}{none}}
	{
		\blockmatrix{#2}{#3}{#4}{black}{none}{true}{black}{none}{#5}
	}
	{
		\definecolor{fillcolor}{rgb}{#1}
		\blockmatrix{#2}{#3}{#4}{black}{none}{true}{black}{fillcolor}{#5}
	}
	
	\end{tikzpicture}
}

\newcommand{\valignbox}[1]{
	\vtop{\null\hbox{#1}}
}

\newenvironment{blockmatrixtabular}
{
	\begin{tabular}{
			@{}c@{}c@{}c@{}c@{}c@{}c@{}c@{}l@{}l@{}l@{}l@{}l@{}l@{}l@{}l@{}l@{}l@{}l@{}l
			@{}c@{}c@{}c@{}l@{}l@{}l@{}l@{}l@{}l@{}l@{}l@{}l@{}l@{}l@{}l@{}l@{}l@{}l@{}l
			@{}l@{}l@{}l@{}l@{}l@{}l@{}l@{}l@{}l@{}l@{}l@{}l@{}l@{}l@{}l@{}l@{}l@{}l@{}l
			@{}
		}
	}
	{
	\end{tabular}
}

\usepackage{setspace}

\newcommand{\fofr}{\theta} 
\newcommand{\fofh}{\eta}
\newcommand{\act}{\phi} 
\newcommand{\ploss}{{\cL}} 
\newcommand{\jloss}{\bar{\cL}} 
\newcommand{\mat}[2]{M\prn*{#1,#2}}

\setlist[itemize]{noitemsep, topsep=0pt}
\setlist[enumerate]{noitemsep, topsep=0pt,label=\arabic*.}

\title{On the Principle of Least Symmetry Breaking\\ in Shallow ReLU Models}

\begin{document}

\author{
	Yossi Arjevani \\
	NYU\\
	\texttt{yossi.arjevani@gmail.com} \\
	\and
	Michael Field\\                                 %
	UCSB\\
	\texttt{mikefield@gmail.com}\\
}
\date{}
\maketitle
\begin{abstract}
We consider the optimization problem associated with 
fitting two-layer ReLU networks with respect to the squared loss, where 
labels are assumed to be generated by a target network. Focusing first on 
standard Gaussian inputs, we show that the structure of spurious 
local minima detected by stochastic gradient descent (SGD) is, in a 
well-defined sense, the \emph{least loss of symmetry} 
with respect to the target weights. 
A~closer 
look at the analysis indicates that this 
principle of least symmetry breaking may apply to a broader range of 
settings. Motivated by this, we conduct a series of experiments which 
corroborate this hypothesis for different classes of non-isotropic 
non-product distributions, smooth activation functions and networks with a 
few layers.
\end{abstract}

\section{Introduction}
The great empirical success of artificial neural networks over the 
past few years has challenged the very foundations of our understanding of 
statistical learning processes. Although fitting generic high-dimensional 
nonconvex models is typicality a lost cause, multilayered ReLU networks 
achieve state-of-the-art 
performance in many machine learning tasks, while being trained using 
numerical solvers as simple as stochastic first-order methods. Given 
this perplexing state of affairs, a large body of recent works has focused 
on various two-layers 
networks as a more manageable means of investigating some of the complexity 
exhibited by deep models. One such major line of research has been 
concerned with exploring the set of assumptions under which the loss 
surface of Gaussian inputs qualifies convergence of various 
local search methods (e.g., 		
\cite{brutzkus2017globally,du2017gradient,li2017convergence,feizi2017porcupine,
zhang2017electron,ge2017learning,tian2017analytical}). Recently, 
Safran \& Shamir 
\cite{safran2017spurious} considered the well-studied 
two-layers ReLU network $\x \mapsto \bones^\top_k\phi(\W\x)$, where $\x \in 
\RR^d, \W\in\mat{k}{d}$, $\phi(z)\defoo \max\{0,z\}$ and 
$\bones_k$ is the $k$-dimensional vector of all ones, and proved that the 
expected squared loss w.r.t. a target network with identity weight matrix, 
i.e., 
\begin{align} \label{def:ploss}
 \ploss(\W) \defoo 
	\frac{1}{2}\EE_{\x\sim\cN(\bzeros,\I_d)}\brk*{ 
		\prn{	 \bones^\top_k\phi(\W\x)- \bones^\top_d\phi(\x) }^2  
		},\quad \W\in\mat{k}{d},
\end{align}
possesses a large number of spurious local minima which can cause 
gradient-based methods to fail. \\

Unlike the existence of multiple spurious local minima, the fact that  
$\ploss$ contains many \emph{global} minima should come as no 
surprise---it is a consequence of symmetry properties of $\ploss$. %
Indeed, assuming for the moment that $k=d$, 
$\ploss$ is 
invariant under all transformations of the form $\W\mapsto \P_\pi 
\W\P_\rho^\top, 
\pi,\rho\in S_d$, where $S_d$ is the symmetric group of degree $d$, and 
$\P_\pi,\P_\rho$ are the 
associated permutation matrices which permute the rows and the columns of 
$\W$ respectively (see \pref{sec:main_res} for more  details). In 
particular, since the 
identity matrix is a global minimizer, so is every permutation 
matrix. 
Permutation matrices are highly `symmetric' in the sense that  
	the group of all row and column permutations that fix a permutation 
	matrix $\P$ --- the \emph{isotropy group} of $\P$ --- is isomorphic to
$\Delta {S_d} \defoo \{\W\mapsto \P_\pi \W  
\P_\pi^\top~|~\pi\in 
S_d\}$ 	(this is  obvious for $\W=\I_d$, since $\I_d = \P_\pi \I_d 
\P_\pi^\top$ holds trivially for any  $\pi\in S_d$).	
With this in mind, one may ask
\begin{center}
\emph{How symmetric are local minima, which are not global}? 	 	
\end{center}
\newcommand{\ibwidth}{1.13in}
\newcommand{\ibheight}{\ibwidth}
\newcommand{\idwidth}{0.1in}
\newcommand{\mtwidth}{0.75}
\newcommand{\mbwidth}{0.23}
\begin{center}
\begin{minipage}{\mtwidth \textwidth}
	In this work, we show that, perhaps surprisingly, the 
	isotropy subgroups of spurious local minima of $\ploss$ are maximal (or 
	large) in $\Delta S_d$, the 
	isotropy of 
	the global solution $\W=\I_d$. In other words,  
	spurious local minima exhibit 
	\emph{the least symmetry loss w.r.t. the symmetry of the target model}. 
	This provides a natural explanation for the patterns that appear in the 
	local minima found in \cite{safran2017spurious} (see  
	\pref{fig:intro_ex}). Next, in examining the 
	predictive power of such a
	principle, we ask how does the 	symmetry of local 
	minima change if $d$ is even and  
	the target model is of reduced symmetry, say, $\W=\I_{d/2}\oplus 
	2\I_{d/2}$ whose isotropy is $\Delta S_{{d}/{2}} \times \Delta 
	S_{{d}/{2}}$? Similarly, we find that maximal isotropy subgroups 
	correctly predict the shape of critical points detected by SGD. A more 
	detailed  analysis reveals that the invariance properties of $\ploss$ 
	do not directly depend on the rotational invariance of standard 
	Gaussian inputs. Indeed, the same structure of critical points is 
	observed when the input is drawn uniformly at random from $[-1,1]^d$, 
	suggesting that isotropy type of critical points strongly depend on the 
	intricate interplay between the network architecture, the input 
	distribution and the label distribution (cf. 
	\cite{brutzkus2017globally,shamir2018distribution} and references 
	therein, for hardness results of learnability under partial sets of 
	assumptions). \\
			
	It is noteworthy that this phenomenon,	in which critical points of 
	invariant functions exhibit maximal (or 
	large) isotropy groups, has been observed many times in various 
	scientific fields;	for example in
	Higgs-Landau theory, equivariant bifurcation theory and replica 
	symmetry breaking (see, 
	e.g.,\cite{Michel1980,Golubitsky1983,field1987symmetry,DingSS15}).
	It appears that ReLU networks form a rather unexpected instance for 
	this 
 principle.  
\end{minipage}
\begin{minipage}{0.2\textwidth}
\end{minipage}
\begin{minipage}{\mbwidth\textwidth}
\begin{figure}[H] 
	\begin{center}
		\includegraphics[scale=0.5]{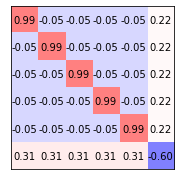}		
		\\		\vskip0.15cm
	\hskip0.1cm	\begin{blockmatrixtabular}
				\dblockmatrixSDMO[1.0,0.8,0.8]{\ibwidth}{\ibheight}{$\alpha$}{\idwidth}{$\beta$}{0.18in}
		\end{blockmatrixtabular}
	\end{center}
	\vskip-0.5cm
	\caption{Top: a $6\times 6$ spurious minimum of 
	(\ref{def:ploss})  \cite[Example 
	1]{safran2017spurious}.
		Bottom: maximal isotropy $\Delta S_5 \times 
		\Delta S_1$ in $\Delta S_6$, the isotropy of the global solution 
		$\I_6$ $(\alpha,\beta,\gamma,\delta,\epsilon\in\RR)$.}
\end{figure}\label{fig:intro_ex}
\end{minipage}
\end{center}
Our contributions (in order of appearance) can be stated as follows: 
\begin{itemize}[topsep=4pt,itemsep=5pt, leftmargin=* ]
	\item We express the invariance properties of ReLU networks in terms of
	the network architecture, as well as the underlying data distribution. 
	As a simple	application, we show that the \emph{auto-balancing 
	property} introduced in \cite{du2018algorithmic} 
	can be reformulated as a \emph{conservation law} induced by 
	a suitable continuous symmetry. This provides an alternative and 
	arguably more natural derivation of this property.

	\item We analyze the intricate isotropy subgroup lattice of $S_k\times 
	S_d$ and $\Delta S_d$, and characterize large classes of the respective 
	maximal isotropy subgroups. Our purely theoretical analysis gives a 
	precise description of 	the structure of spurious local minima detected 
	by SGD for 	ReLU network (\ref{def:ploss}). We show that this is 
	consistent over various target weight matrices.

	\item We conduct a series of experiments which examine 
	how the isotropy 	types of critical points depend on the \emph{joint 
	symmetry} of the 
	target model, the underlying distribution and the 
	network architecture. 
	Our findings indicate that a 
	similar phenomenon indeed applies to other related problems with, 
	e.g., Leaky-ReLU and softplus activation functions, slightly 
	over-specified parameterization and a 
	few fully-connected layers. We also demonstrate some 
	settings where 
	this phenomenon does not occur. 
\end{itemize}

The paper is organized as follows. After surveying related work below, and 
relevant group-theoretic background in \pref{sec:preliminaries}, we 
present a simple derivation of the auto-balancing property based on 
symmetry arguments. In \pref{sec:main_res} we provide a 
detailed analysis of the invariance properties of ReLU networks and the 
corresponding isotropy lattice. Lastly, we present in 
\pref{sec:empirical_res} empirical results which examine the 
scope of the least (or small) symmetry breaking principle. All proofs are 
deferred 
to the appendix.

\section{Related Works}

Two-layer ReLU networks, the main focus of this work, have been studied  
under various settings which differ by their choice of, for example, activation 
function, underlying data distribution, number of hidden layers w.r.t. 
number of samples and numerical solvers
(\cite{brutzkus2017sgd,li2018learning,soltanolkotabi2018theoretical,xie2016diverse,
zhong2017recovery,panigrahy2017convergence,		
du2018gradient,janzamin2015beating}).  Closer to our settings
are works which consider Gaussian inputs and 
related obstacles for optimization methods, such as bad local minima. 
\cite{zhang2017electron,du2017gradient,feizi2017porcupine,
li2017convergence,tian2017analytical,brutzkus2017globally,ge2017learning}. 
Most notably, the spurious minima addressed by \cite{safran2017spurious} 
exhibit the very symmetry claimed in this work. Another relevant line of 
works have studied trainability of ReLU multilayered networks directly 
through the lens of symmetry. Concretely, it is shown that the rich 
weight-space symmetry carries important information on singularities which 
can significantly harm the convergence of gradient-based methods 	
\cite{saad1995line,amari2006singularities,wei2008dynamics,orhan2017skip,brea2019weight}.
In contrast to this, symmetry here is utilized as a mean of studying the 
structure of critical points, rather than characterizing various regions of 
the loss landscape.

\paragraph{Symmetry breaking in nonconvex optimization}
The present work has been the precursor to a rapidly growing body of work building on the phenomena of symmetry breaking first reported and studied here. In 
\cite{arjevani2021symmetry}, path-based techniques are introduced, 
allowing the construction of infinite families of critical points 
for ReLU two-layer networks using Puiseux series. In 
\cite{arjevani2020analytic}, results from the representation 
theory of the symmetric group are used to obtain precise analytic 
estimates on the Hessian spectrum. In~\cite{arjevani2021analytic}, it is 
shown that certain families of saddles transform into spurious minima 
at a fractional dimensionality. In addition, Hessian spectra at 
spurious minima are shown to coincide with that of global minima 
modulo $O(d^{-{1}/{2}})$-terms. In 
\cite{arjevani2022annihilation}, it is proved that adding 
neurons can turn symmetric spurious minima into saddles. In 
\cite{arjevani2022equivariant}, generic $S_d$-equivariant 
steady-state bifurcation is studied, emphasizing irreducible representations along which 
spurious minima may be created and annihilated. In \cite{arjevani2023hidden}, it is shown that the way subspaces \emph{invariant} to the action of subgroups of $S_d$ are arranged relative to ones \emph{fixed} by the action determines the admissible types of structure and symmetry of curves along which $\ploss$ is minimized and maximized. Symmetry breaking phenomena were also shown to occur for tensor decomposition problems  \cite{arjevani2021symmetrytensor}, and later used for the derivation of Puiseux series of families of critical points and the construction of tangency arcs relative to third-order saddles \cite{arjevani2023symmetry}.

\section{Preliminaries}\label{sec:preliminaries}
We briefly review some relevant	background material on group actions, 
representations and equivariant maps  and fix relevant notations (for a 
more complete account see \cite[Chapters 1, 2]{Field2007}). 
Elementary notions and properties in group theory are assumed known. We 
begin by listing a few notable examples which will be 
referred to in later sections.
\begin{exams} \label{exams:groups}
\begin{enumerate}[leftmargin=*,noitemsep]
	\item We let $\RR^*$ denote the group of positive real numbers 
	with multiplication. 
	\item For any finite dimensional vector space $E$, let $\GLn{E}$ denote 
	the \emph{general linear group} of invertible linear transformations of 
	$E$ (in particular, $\RR^*$ is a subgroup of $\GLn{\real}$). Note that 
	if we take 	the standard Euclidean basis of $\RR^d$, then there is 
	a natural 
	isomorphism between $\GLn{\RR^d}$ and $\GLN$, the group of all 
	invertible $d \times d$ real matrices, which is often regarded as 
	an identification.  If we take the standard Eucliean inner product on 
	$\real^d$, with the norm $\|\;\|$, then 
	\[
	\ON = 	\{A \in \GLN \dd \|A\x\| = \|\x\|,\;\text{for all } \x \in 
	\real^d\},
	\] 
	is the group of \emph{orthogonal matrices}. In particular, $\ON = \{A 
	\in \GLN \dd A^{-1} = A^\top\}$. Both $\GLN$ and $\ON$ are examples of 
	Lie groups --- groups that have the 
	structure of a smooth manifold w.r.t. which
	the group operations are smooth maps. 
	\item The symmetric group $S_d$ of all permutations of $[d]\defoo 
	\{1,\dots,d\}$ is of special interest for us.  The group $S_d$ can be 
	identified 
	with the subgroup of permutation matrices of $\ON$ (see below).
	Similarly, the hyperoctahedral group $H_d = \intg_2 \wr S_d$ of 
	symmetries of the unit hypercube in $\real^d$
	can be identified with the subgroup of \emph{signed} permutation matrices of $\ON$.

\end{enumerate}
\end{exams}
Characteristically, the groups described in (2,3) above consist of 
transformations of an underlying set. This leads naturally to the 
concept of a \emph{group action} on a set $X$, i.e., a 
group homomorphism from $G$ to the group of bijections of $X$. 
For instance, the groups $\GLN$ and 
$\ON$ naturally act on $\RR^d$ as linear maps (or via standard 
matrix-vector multiplication). Another example, which we use extensively in 
studying the invariance properties of ReLU networks, is the action of the 
group $\Gamma_{k,d} = S_k \times S_d \subseteq S_{k+d},~k,d\in \NN$ on 
$[k] 	\times [d]$ defined by 
\begin{align}\label{eq: Gamma-action}
	(\pi,\rho)(i,j) = (\pi^{-1}(i),\rho^{-1}(j)),\; \pi \in S_k, 
\rho \in S_d,\; (i,j) \in [k] \times [d].
\end{align}

Given a $G$-space $X$ and $x \in X$, we define 
$Gx = \{gx \dd g \in G\}$ to be the \emph{$G$-orbit} of $x$, and 
$G_x = \{g \in G \dd gx = x\}$ to be the \emph{isotropy} 
subgroup of $G$ at $x$.	Subgroups $H, H'$ of $G$ are  
\emph{conjugate} if there exists $g \in G$ such that $gHg^{-1} = H'$.  
Points $x, x' \in X$ have the same \emph{isotropy type} (or same symmetry) 
if $G_x, G_{x'}$ are conjugate subgroups of $G$. Note that, points on 
the same $G$-orbit have the same isotropy type since 
$G_{gx} = gG_x g^{-1}$, for all $g \in G$. The action of $G$ is 
\emph{transitive} if for any $x,y \in X$ there exists $g\in G$ such that  
$gx = y$, and \emph{doubly transitive} if	
for all $x,x',y,y' \in 	G$, $x \ne x'$, $y \ne y'$, there exists $g \in 
G$ such that $gx = y$, $gx' = y'$. Lastly, the \emph{transitivity 
partition} of subgroup $H\subseteq G$ is the set of all $H$-orbits 
$\{Hx~|~x\in X\}$. \\  

We are mainly interested in \emph{linear} actions, equivalently 
\emph{representations}, of $G$ which are given by a homomorphism $R: G \arr 
\GLn{E}$, where $E$ is a vector space. 
For example, $R_1(t) = t$, $ t \in \RR^*$ defines a representation of 
$\RR^*$ in $\GLn{1}$ and if $R_2(t)$ is the $2\times 2$-diagonal matrix 
with entries
$t, t^{-1}$, then $R_2$ defines a representation of $\RR^*$ in 
$\GLn{2}$.  
   	The groups $\GLN$ and $\ON$ are naturally  represented on $\RR^d$ through the inclusion map. The group $S_d$ can 
   	be represented on $\RR^d$ by associating each $\pi\in S_d$ with the 
   	$d\times d$-permutation 
matrix
\begin{align} \label{def:permutation_matrix}
	(\P_\pi)_{ij} = 
	\begin{cases}
		1 & i = \pi(j),\\
		0 & \text{o.w.}.
	\end{cases}
\end{align}
Since $S_k$, $S_d$ can be identified with groups of permutation matrices, 
we have a representation of	$\Gamma_{k,d}$ which associate a pair of 
permutations $(\pi,\rho)$ with the linear transformation $\W \mapsto 
\P_{\pi}\W\P_{\rho}^\top$ (equivalently, $ 
(\pi,\rho)[\W_{ij}] = [\W_{\pi^{-1}(i),\rho^{-1}(j)}]$).
Slightly abusing notation, we shall occasionally refer the permutation 
$(\pi,\rho)$ of $S_k \times S_d$ as a transformation  
of $M(k,d)$. 
Similarly, the group $\HN$ admits a representation in $\GLN$ as the
group of \emph{signed permutation matrices} --- permutation matrices with 
entries $\pm 1$. 
We note that the representations described above are the only 
representations we use. Given two representations of a group $G$ on vector 
spaces $E_1$ and $E_2$, a map $F: E_1 \arr E_2$ is \emph{$G$-equivariant} 
if $F(g\x) = g 	F(\x),\; \x\in E_1,\; g \in G$. A map $f:E_1 \arr \RR$ is
$G$-\emph{invariant} if $f(g\x) = f(\x),\; \x \in X, \; g \in G$.
\begin{exams} \label{ex:invariant_functions}
	\begin{enumerate}[leftmargin=*]
		\item The squared Euclidean norm $\|\cdot \|^2$ on $\RR^d$ is 
		$G$-invariant for all $G \subset \ON$. 
		\item The function $\fofr(x,y)\defoo x^2y^2$ 
		on $\RR^2$ is $R^*$-invariant w.r.t. representation $R_2$
		defined above. \label{expr:r_func} 
		\item The function $\fofh(\x) \defoo  \nrm{\x}^4 + 
		\sum^n_{i=1} {x_i^4} - \nrm{\x}^2$ on $\RR^d$ is 
		$\HN$-invariant.\label{expr:h_func}
	\end{enumerate}
\end{exams}
In the context of this work, one key feature of $G$-invariant 
differentiable functions is the $G$-equivariance of their gradient fields 
(and other higher-order derivatives). These are naturally expressed 
in terms of fixed point linear subspaces  
defined by $X^H \defoo \{y \in X \dd hy =y, \forall h\in H\},~ H\subseteq 
G$. 
\begin{prop}\label{prop: grad}
	If $G$ is a closed subgroup of $\On{E}$ and $f:E \arr \real$ is  
	$G$-invariant, then the gradient vector field of $f$, $\grad{f}:E 
	\arr E$, is  $G$-equivariant. In particular, if $H$ is a subgroup of 
	$G$, then
	\begin{enumerate}[topsep=0pt]
		\item $\grad{(f|\Omega^H)} = \grad{f} | \Omega^H$. Thus, 
		$\c \in \Omega^H$ is a critical point of $\grad{(f|\Omega^H) }
		$ iff $\c$ is a critical point of $\grad{f}$.
		\label{item:tangent} 
		\item  Gradient flows which are initialized in $E^H$ must stay in 
		${E^H}$ (`isotropy can only grow over time').
		\item Eigenvalues of $\hes{(f|\Omega^H)} $ determine the 
		subset of eigenvalues of $\hes{f} | \Omega^H$
		associated to directions tangent to $E^H$ (nothing is said about 
		directions transverse to $E^H$).
\end{enumerate}
\end{prop}
Note that this immediately implies that both gradient maps 
	$\grad({\|\x\|^2)}$ and $\grad{\fofh}(\x)$ (see 
	\pref{ex:invariant_functions}.\ref{expr:h_func} above)
are $\ON$- and $\HN$-equivariant, respectively. In 
addition, a simple, yet 
useful, implication of \pref{prop: grad}, is that when 
the sublevel 
sets of a given invariant function are compact, 
\emph{any} fixed subspace 
$X^H$ contains at least  one critical point. In other 
words, for any 
$H\subseteq G$, there exists a critical point which 
remain fixed under the 
action of $H$.\\ 

We now present a simple application of the group-theoretic 
framework introduced above. The application concerns quantities conserved 
by gradient flows of invariant functions (which in turn implies 
\emph{algorithmic regularization} of the vanilla gradient descent 
algorithm). Consider the $R_2$-invariant function $\fofr$ 
introduced in \pref{ex:invariant_functions}.\ref{expr:r_func}. Since $\fofr 
= \fofr \circ R_2(t),~t\in\RR^*$, it follows that for any 
$(x,y)\in\RR^2$
\begin{align*}
	0 = \frac{d \fofr(x,y)}{d t} = \frac{d \fofr( 	
	R_2(t) (x,y)) }{d t} = D\fofr (R_2(t)(x,y)) 
	\mymat{1&0\\0&-t^{-2}}\mymat{x\\y}.
\end{align*}
In particular for $t=1$, we get $(x,-y)\nabla \fofr(x,y)=0$. Therefore, any 
gradient flow $\gamma(s)$ of $\fofr$ must satisfy $(\gamma_1(s), 
-\gamma_2(s)) \gamma'(s)=0$. Setting $\Phi(x,y)=(x^2-y^2)/2$, this can be 
equivalently expressed as $d\Phi(\gamma(s))/ds = 0$. Hence, for any $s>0$, 
it holds that $\Phi(\gamma(s))=\Phi(\gamma(0))$, implying that $\Phi$ is 
conserved by $\gamma(s)$. 
This simple example directly generalizes to ReLU networks (see 
\pref{def:multi_relu} below). Let $E_1,\dots, E_N$ be vector spaces 
and let $f$ be a real-valued function defined over $E_1\times\cdots\times 
E_N$ such that 	\begin{align} \label{def:r_star_inv}
f(\v^{(1)},\dots,\v^{(N)})=f(\v^{(1)},\dots,c\v^{(i)}\dots,(1/c)\v^{(j)},\dots,\v^{(N)}),~\v^{(i)}\in
E_i.
\end{align}
Then, the same argument shows that 
%
\begin{align} \label{eqn:consrstarinv}
\Phi(\v^{(1)},\dots,\v^{(N)})= 	(\|\v^{(i)}\|^2-\|\v^{(j)}\|^2)/2 
\end{align}
must be conserved by any gradient flow of $f$. Since ReLU activation 
functions are positively homogeneous, this gives a succinct and perhaps 
more intuitive symmetry-based derivation of the auto-balancing property 
considered in \cite[Theorem 2.1]{du2018algorithmic}. Also note that this 
derivation applies to any function satisfying invariance properties 
(\ref{def:r_star_inv}) --- not just ReLU networks.	Likewise, if 	
$E_i=M(d_{i},d_{i-1})$ and $f$ 
satisfies 
\begin{align}	\label{def:gln_inv}
f(\W^{(1)},\dots,\W^{(N)})=f(\W^{(1)},\dots,  \A \W^{(i)}, 
\W^{(i+1)} 	\A^{-1},\dots,\W^{(N)}),~\W^{(i)}\in E_i,
\end{align}
for any $\A\in \GLn{M(d_{i},d_{i})}$, then a similar argument shows that 
\begin{align}\label{eqn:glninv}
\Phi(\W^{(1)},\dots,\W^{(N)})= 
\prn*{\W^{(i)}(\W^{(i)})^\top- 	
	(\W^{(i+1)})^\top \W^{(i+1)}}/2
\end{align} 
is conserved by any gradient flow (see \pref{sec:eqn_glninv}). This 
holds in particular for 
neural networks with linear activation (see \cite[Theorem 
2.2]{du2018algorithmic}). \\

Whereas continuous symmetry groups are related to the  
dynamics of gradient flows and gradient-based optimization 
methods, it appears that in our setting, the most   
important ingredients in studying the structure of critical 
points are discrete symmetry groups. We devote the remainder 
of the paper to this topic. 

\section{Isotropy Types of ReLU Neural Networks} \label{sec:main_res}
As mentioned earlier, the notion of isotropy groups provides a natural 
means of measuring the 
symmetry of points relative to a $G$-action --- the more symmetric the  
point, the larger the isotropy group. The most symmetric points are 
those with isotropy group $G$, and the least symmetric are points with 
trivial isotropy group containing only the identity element. This 
naturally raises the question: \emph{How symmetric are critical 
and extremal points of invariant functions?} In the realm of convexity, it 	
is a nearly-trivial fact that critical points of strictly convex invariant 
functions must be of full isotropy 	type\todoc{ref}. For nonconvex 
invariant functions, matters are anything 
but trivial. For example, a straightforward analysis shows that the 
$\HN$-invariant function $\fofh$, defined in 
\pref{ex:invariant_functions}.\ref{expr:h_func}, has
$3^n$ critical points  specified as follows (see 
\pref{sec:h_analysis} for 
a full derivation): For	any $p\in\{0,\dots,d\}$, $\fofh$ possesses 
$\binom{d}{p}2^p$ critical points	of isotropy type $S_p\times H_{d-p}$ 
--- all of which are maximal (w.r.t. set inclusion) proper isotropy  
subgroups of $\HN$!	
In what follows, we show that a similar phenomenon is exhibited by ReLU 
networks. We start with a detailed examination of the simple ReLU instance 
(\ref{def:ploss}), and then briefly cover some related generalizations 
towards the end of the section. 

\subsection{\texorpdfstring{$\Gamma_{k,d}$-invariance of 
$\ploss$}{Invariance Properties of the Loss 
Function}}\label{sec:gamma_inv}
We start by showing that $\ploss$ (\pref{def:ploss}) is 
$\Gamma_{k,d}$-invariant. First, we make the dependence of $\ploss$ on the 
target weight $d\times d$-matrix 
$\V$ explicit:
\begin{align} \label{def:jloss}
	\jloss(\W,\V) &\defoo \frac12 \E_{\x\sim 
	\cN(\bzeros,\I_d)}\brk*{\prn*{  \bones_k^\top \act(\W\x)  - 
	\bones_d^\top 
	\act(\V\x)     
	}^2}. 
\end{align}
Next, observe that for any $\pi\in S_k, \rho \in S_d$ and 
$\U\in\On{d}$, we have
\begin{align}
	\jloss(\W,\V) &= \jloss(\P_\pi \W,\V) = 
	\jloss(\W,\P_\rho \V),\label{property_one}\\
	\jloss(\W,\V)&= \jloss(\W \U,\V\U),\label{property_two}
\end{align}
where the last equality follows by the $\ON$-invariance of the standard 
multivariate Gaussian distribution. Therefore, for any $\rho \in S_d$ and 
$\U\in\ON$ such that $\V = \P_\rho \V\U^\top$ and any $\pi \in S_k$, we have
\begin{align*} 
\jloss(\W,\V) = \jloss(\W,\P_\rho \V\U^\top) 
\stackrel{(\ref{property_one})}{=} 
\jloss(\W,\V\U^\top) \stackrel{(\ref{property_two})}{=}  \jloss(\W\U , 
\V\U^\top \U) =  \jloss(\W\U , \V)\stackrel{(\ref{property_one})}{=} 
\jloss(\P_\pi \W\U , \V).
\end{align*}
In particular, for $\V=\I_{d}$, we have $\V=\P_\pi \V\P_\pi^\top$ for any 
$\pi \in S_d$, from which it follows that $\ploss(\W)=\jloss(\W,\I_d)$ is  
$\Gamma_{k,d}$-invariant w.r.t. $\W$. Note that here we do not exploit the 
rotation invariance of the standard Gaussian distribution, but  
rather its invariance under permutations. Hence, the same 
$\Gamma_{k,d}$-invariance holds for any product distribution when 
$\V=\I_d$. 
For example, in	\pref{sec:empirical_res} we show that critical points also 
admit maximal isotropy types when the input distribution is 
$\cD = \cU({[-1,1]^d})$ (but not when $\cD = \cU({[0,2]^d})$).

\subsection{Isotropy Subgroups of \texorpdfstring{$\Gamma_{d,d}$}{Gamma}}
Having established the $\Gamma_{k,d}$-invariance of $\cL$, our next goal is 
to consider the corresponding lattice of isotropy subgroups.
After some preliminary generalities and examples, we focus on isotropy 
subgroups of \emph{diagonal type}, that is, groups 
which are conjugate to subgroups of $\Delta S_d \defoo 	\{(\pi 
,\pi)~|~\pi \in S_d\}\subseteq S_d^2$, 	the isotropy of the target weight 
matrix $\I_d$. Note that when the 
underlying distribution is unitary invariant, the same analysis applies 
to any orthonormal weight matrix, mutatis mutandis. \\

Given a subgroup $H$ of $S_p$, $p > 1$, let $\cP = 
\{P_1,\cdots,P_s\}$ 
be the	$H$-transitivity partition of $\is{p}$ (in 
particular, $H$ 
acts transitively on each  $P_j$, $j \in \is{s}$). 
After a relabeling 
of 	$\is{p}$, we may assume that $P_1 =	\{1,\ldots, 
p_1\}$, $P_2 = \{p_1 
+1 , 
\ldots, p_2\}$, $\ldots$, $P_s = \{p_{s-1}+1,\ldots,p_s = d\}$, where $1 
\le p_1 < p_2 < \cdots < p_s=d$.  A
partition $\cP$ satisfying this condition is \emph{normalized}. 	
Suppose that $H$ is a subgroup of $\Gamma$. For $j = 
1,2$, let $H_j = \pi_j H \subset S_d$ denote the projection of $H$ onto the
$j$th factor of $S_d \times S_d$. Note our convention that the group $H_1$ 
permutes rows, $H_2$ permutes columns.
Let $\cP = \{P_a\dd a \in \is{p} \}$, and $\cQ= \{Q_b\dd b \in 
\is{q}\}$ respectively denote the transitivity partitions of the actions of 
$H_1$ and $H_2$ on $\is{d}$	and assume $\cP, \cQ$ are 
normalized. 
Each rectangle $R_{ab} = P_a \times Q_b$, $a \in \is{p}$, $b \in \is{q}$, 
is $H$-invariant (in $\is{d}^2$) and $H$ acts transitively on the rows and 
columns of $R_{ab}$. We refer to the collection $\cR = \{R_{ab} \dd a \in 
\is{p}, b \in \is{q}\}$ as the partition of $\is{d}^2$ by rectangles.  Note 
that the partition $\mathcal{R}$ is maximal: any non-empty $H$-invariant 
rectangle contained in $R_{ab} \in \mathcal{R}$ must equal  $R_{ab}$. 		
In general, $H$ does \emph{not} act transitively on the rectangles 	
$R_{ab}\in \mathcal{R}$. For example, take $H = \Delta S_d$, $d > 1$.		
We have $H_1, H_2 = S_d$ and the transitivity partition for the action 		
of $\Delta S_d$ on $\is{d}^2$ has two parts: the diagonal $\{(i,i) \dd 		
i \in \is{d}\}$ and its complement $\{(i,j) \dd i,j \in \is{d}, i \ne 		
j\}$. \\

The arguments above allow us to reduce the analysis of $H$-actions on 
$\is{d}^2$ to the study of the $H$-action on individual rectangles of 
$\mathcal{R}$. Fixing $R_{ab}\in\mathcal{R}$, let 	$\cT^{ab} = \cT =  
\{T^{ab}_i\dd i \in \is{t_{ab}}\}$ denote the 	transitivity partition for 
the 
action of $H$ on $R_{ab}$. If $\W \in \mat{k}{k}$, let $\cR^\W = 
\{R_{ab}^\W \dd a \in \is{p}, b \in \is{q}\}$ denote the decomposition of 
$\W$ into the submatrices of $\W$ induced by $\cR$. 
The rectangles $R_{ab}$ share the following  useful
property.
\begin{lemma}\label{lem:rect}
	Let $H$ be a subgroup of $\Gamma$ with associated partition $\cR$ of 
	$[d]^2$ by rectangles, and let $(a,b)\in [p]\times [q]$. 
	For all $\ell \in \is{t_{ab}}$, each row and column of $R_{ab}$ 
	contains 
	the same number of elements of $T^{ab}_\ell$. In particular, if $\W \in 
	\mat{d}{d}$ and $\Gamma_\W = H$, then row sums and column sums 
	for the submatrix $R_{ab}^\W$ are equal. 
\end{lemma}
If the isotropy group is a product of subgroups of $S_d$, then 
$R_{ab}$ takes a particularly simple form.  
\begin{lemma}\label{lem:prod}
	Suppose $H = H_1 \times H_2\subset S_d \times S_d$ and $H= 
	\Gamma_\W$. Then each rectangle $R_{ab}^\W$ will have all entries 
	equal and, if $\cP, \cQ$ are normalized, 
	$\Gamma_\W = (\prod_{j=1}^p S_{a_j}) \times (\prod_{i=1}^q S_{b_i})$,  
	where $S_{a_j} \times S_{b_i}$ fixes $R_{a_jb_i}^\W$. 
\end{lemma}
Note, however, that while matrices with product isotropy groups have a	
simple structure, matters are not always so simple. For 
example, for any distinct $\alpha, \beta, \gamma,\delta $, the 	
respective isotropy groups of the $4\times 4$ matrices 
\begin{align} \label{ex:prod_not_prod}
	\begin{array}{ccccccc}
	 \left[\begin{matrix}
		\alpha & \alpha & \alpha & \alpha 
		\\                                   
		\alpha & \alpha & \alpha & \alpha 
		\\                                   
		\beta  & \beta  & \beta  & \beta  \\
		\beta  & \beta  & \beta  & \beta                                    
		\end{matrix}\right],&
		\quad\quad
  \left[\begin{matrix}
		\alpha & \beta & \beta & \beta \\                                   
		\alpha & \beta & \beta & \beta \\                                   
		\gamma & \delta & \delta & \delta \\
		\gamma & \delta & \delta & 
		\delta                                    
		\end{matrix}\right],&
		\quad \quad
     \left[\begin{matrix}
		\alpha & \alpha & \beta & \beta 
		\\                                   
		\beta & \beta & \alpha & \alpha 
		\\                                   
		\alpha & \beta & \alpha & \beta \\
		\beta & \alpha & \beta & \alpha                                    
		\end{matrix}\right],
	\end{array}	
\end{align}
are the product groups 	$K_1 \defoo \inner{(12),(34)}\times S_4$,~$K_2 \defoo \inner{(12),(34)}\times (\{(1)\}\times S_3)$ and a 
group $K_3$ which is \emph{not} isomorphic to a product (but rather to the 	
group $H_2$ of order $8$. See  \pref{sec:not_diag_app} for more 
details). Clearly, the latter induces a relatively more complex 
structure. We remark that fixed subspaces of  
isotropy groups carry important information regarding the 
dynamics of 
gradient flows and gradient-based algorithms. By 
\pref{prop: grad}, any trajectory initialized in say $M(4,4,)^{K_1}$, (or 
close to it if 
the subspace is transversally stable) always stays 
close to the subspace (up to numerical stability). Likewise, when the 
weight vectors 
assigned to two 
different hidden neurons are identical (equivalently, if 
two rows of the 
weight 
matrix are identical), they must remain so throughout the optimization via 
gradient-based methods.
Note that this holds regardless of the underlying 
distribution as $\ploss$ is always $S_d$-invariant by the left 
$S_d$-action.\\ 

Unlike the groups $K_i$ discussed above, some subgroups 
cannot be realized as isotropy groups. To show this	groups, we use the 
following lemma.

\begin{lemma} \label{lem:equal_diagonal}
	If $H$ is a transitive subgroup of $S_d$ and $\W\in \mat{d}{d 
	}^{\Delta 
	H}$ (so $\Gamma_\W \supset \Delta H$), then the diagonal elements of 
	$\W$ are all equal.
	Conversely, if the rectangle partition for the action of $\Gamma_\W$ 
	is $\{\is{d}^2\}$ and there exists $(i_0,j_0) \in \is{d}^2$ such that 
	the $\Gamma_\W$-orbit of $(i_0,j_0)$ contains exactly $d$ points, then 
	$\Gamma_\W $ is conjugate to $\Delta K$, where $K$ is a transitive 
	subgroup of $S_d$. 	In particular, $\Gamma_\W  = \Delta S_d$ iff there 
	exists $\alpha,\beta \in \real$, $\alpha \ne \beta$, such that $w_{ii} 
	= \alpha$, $i \in \is{d}$, and $w_{ij} = \beta$, $i,j \in 	\is{d}$, 
	$i \ne j$.
\end{lemma}

If $K \subsetneq S_k$ is a doubly transitive subgroup of $S_d$ (for 
example, the alternating subgroup $A_d$ of $S_d$, $d > 3$), then $\Delta K$ 
is not an isotropy group for the action of $\Gamma$: the double 
transitivity implies that if $\Gamma_\W = \Delta K$ then all off-diagonal 
entries are equal. Hence $\Gamma_\W = \Delta S_d$ by 
\pref{lem:equal_diagonal}. \\

The analysis of isotropy of diagonal type can largely be reduced to the 
study of the diagonal action of transitive subgroups of $S_p$, $2 \le p \le 
d$. We give two examples to illustrate the approach and 
then concentrate on maximal isotropy subgroups of $\Delta S_d$. 
\begin{exams}
	Suppose $H = \intg_4 \subset S_4$ is the cyclic group of order $4$ 
	generated by the $4$-cycle $(1234)$. Matrices with
	isotropy $\Delta \intg_4$ are circulant matrices of the form
	\begin{equation}\label{eq: blockform}
	\W = 
	\left(\begin{matrix}
	\alpha & \beta & \gamma & \delta \\
	\delta & \alpha & \beta & \gamma \\
	\gamma & \delta & \alpha & \beta \\
	\beta & \gamma & \delta & \alpha 
	\end{matrix} \right) 
	\end{equation}
	where $\alpha,\beta,\gamma,\delta$ are distinct (else, the matrix has 
	a  bigger isotropy group). 
\\		(2) If $d = 8$ and $H = \intg_4 \times \intg_4$, then $\Delta H = 
	\Delta \intg_4 \times \Delta \intg_4$. Matrices
	with isotropy $\Delta H$ may be written in block form as
	$
	\W = 
	\left(\begin{matrix}
	A & B \\
	C & D 
	\end{matrix} \right), 
	$
	where the matrices $A,D$ have the structure given by~\Ref{eq: 
	blockform}. Since $\Delta H$ is a product of groups of diagonal type, 
	$B$ and $C$ have all their entries equal. We may vary this example 
	without changing the rectangle partition. For example, if $K \subset 
	S_8$ is  generated by $(1234)(5678)$, then $K \approx \intg_4$. With $H 
	= \Delta K$, we find that a matrix $\W$ with isotropy $\Delta K$ has 
	the same block decomposition as before but now every block has the 
	structure given by~\Ref{eq: blockform} and there will be 16 parameters, 
	4 for each block.
\end{exams}

\subsection{Maximal isotropy subgroups of \texorpdfstring{$\Delta S_d$}{the 
Diagonal Action of the Symmetric Group of Degree d}}
\todoc{We have full $S_k \times S_n$ invariance but we only care symmetry 
	breaking of the diagonal action?}
Of special interest in this work are maximal isotropy subgroups of $\Delta 
S_d = \Gamma_{I_d}$. These subgroups can be conveniently characterized 
through maximal subgroups of $S_d$. Indeed, every subgroup of $\Delta S_d$ 
must be diagonal. Hence, maximal subgroups of $\Delta S_d$ are in 
one-to-one correspondence with the maximal subgroups of 
$S_d$. The latter 
topic has received considerable attention from group 
theorists in part 
because of connections with the classification of simple groups (see 
\cite[Appendix 2]{AschbacherScott1985}). Here we describe two 
relatively-known cases: maximal subgroups of $S_d$ which are 
instransitive (not transitive) and the class of imprimitive transitive
subgroups of $S_d$ (primitive transitive subgroups of $S_d$ are addressed 
in \cite{Liebecketal1987,DixonMortimer1996}).

\begin{lemma} \label{lem:max_subgropus_sn}
	\begin{enumerate} 
		\item If $p+q = d, p \ne q$, then $S_p \times S_{q}$ is a maximal 
		proper subgroup of $S_d$ (intransitive case). \label{item:max1}
		\item If $d = pq$, $p, q > 1$, 	then the wreath product $S_p \wr 
		S_q$ is transitive and a maximal 
		proper subgroup of $S_d$ (imprimitive case). \label{item:max2}
	\end{enumerate}
\end{lemma}
The first category of groups in \pref{lem:max_subgropus_sn} corresponds to
maximal subgroups of $\Delta S_d$ of \emph{hierarchical} nature. Assuming $d
\ge 3$, set $\Delta S_{p,q} = \Delta S_{p}\times \Delta S_q,~p+q=d,~ 0\le 
q<d/2$ (where   $\Delta S_d\times \Delta S_0$ is set to be $\Delta S_d$). 
If 
$\W \in M(d,d)$ has isotropy conjugate to $\Delta S_{p,q}$ then, after a 
permutation of rows and columns, we may write $\W$ in a block matrix form 
$\mymat{A & b \cI_{p,q} \\ c \cI_{q,p}& D }$ where $A\in 
M(p,p)^{\Delta S_{p}}$,
$D\in M(q,q)^{\Delta S_{q}}$, $b,c\in \RR$ and $\cI$ is a matrix with all
entries equal 1 of suitable size. It is straightforward to 
verify that, for 
sufficiently large $d$, 
$\dim M(d,d)^{\Delta S_{p,q}}$ is 2,5 and 6 for $q=0,~q=1$ and $1<q<d/2$, 
respectively (see 
\pref{fig:max_sym}). In particular,  $\dim M(d,d)^{\Delta S_{p,q}}$ 
does not depend on
the ambient dimension $d$ (when sufficiently large).
\newcommand{\bwidth}{1.2in}
\newcommand{\bheight}{\bwidth}
\newcommand{\dwidth}{0.11in}
\begin{figure}[ht]
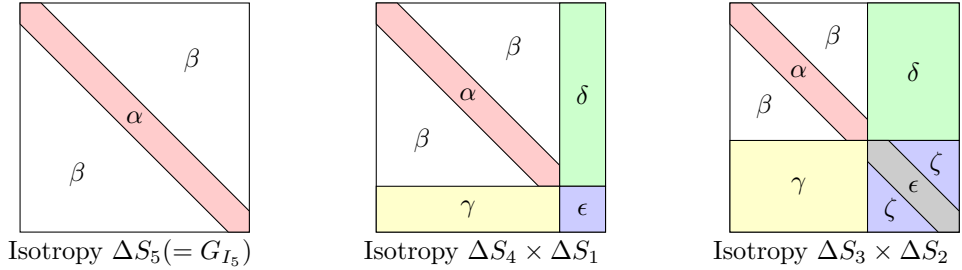
 
\begin{center}\vskip-0.4cm
{\setstretch{1.2}
	\begin{blockmatrixtabular}\label{fig:dia_sym}
	\valignbox{\dblockmatrixSD[1.0,0.8,0.8]{\bwidth}{\bheight}{$\alpha$}{\dwidth}{$\beta$}}&
 \quad\quad\quad\quad
	\valignbox{\dblockmatrixSDMO[1.0,0.8,0.8]{\bwidth}{\bheight}{$\alpha$}{\dwidth}{$\beta$}{0.24in}}&
 \quad\quad\quad\quad
	\valignbox{\dblockmatrixSDMOOO[1.0,0.8,0.8]{\bwidth}{\bheight}{$\alpha$}{\dwidth}{$\beta$}{0.48in}}&\\
	Isotropy $\Delta S_5 (= G_{I_5})$& 		 \quad\quad\quad\quad 
	Isotropy  $\Delta S_4 \times 
	\Delta 
	S_1$  & 		 \quad\quad\quad\quad Isotropy $\Delta 
	S_3\times \Delta
	S_2$
\end{blockmatrixtabular}}
\end{center}
\vskip-0.4cm		
\caption{A schematic description of matrices with isotropy $\Delta 
S_5, \Delta S_4 \times \Delta S_1$ and $\Delta S_3\times S_2$ 
($\alpha,\beta,\gamma,\delta,\epsilon,\zeta\in\RR$ are 
assumed sufficiently `distinctive' so 
as to rule out larger isotropy, 
e.g., $\alpha$ must be different than 
$\beta$).}\label{fig:max_sym}
\end{figure}\\

The second category of maximal subgroups in \pref{lem:max_subgropus_sn} 
induces a relatively more complicated structure. For example, the subgroup 
$K = S_2 \wr S_3$ is a maximal 
transitive subgroup of $S_6$ and has order $48$. Hence $\Delta K$ is a 
maximal subgroup of $\Delta S_6$. Now, if $\Gamma_\W = \Delta K$, then 
\[
\W = 
\left(\begin{matrix}
a & b & c & c & c & c\\
b & a & c & c & c & c\\
c & c & a & b & c & c\\
c & c & b & a & c & c\\
c & c & c & c & a & b\\
c & c & c & c & b & a
\end{matrix} \right) 
\]
and so $\text{dim}(M(6,6)^{\Delta K}) = 3$. More generally, if $K=S_m \wr 
S_n\subseteq S_d$, where $d=mn,~m,n>1$, then $\text{dim}(M(6,6)^{\Delta K}) 
= 3$. We have yet to find critical points of this 
diagonal type category.\\

The analysis above easily extends to multilayered fully-connected 
ReLU networks. Given $d_i\in \NN,~i=0,\dots,N$, with 
$d_0=d$ and $d_N=1$, 
we define 
\begin{align} \label{def:multi_relu}
f(\x;\W^{(1)},\dots, \W^{(N)}) &\defoo \W^{(N)}\phi 
(\W^{(N-1)}\cdots\W^{(2)}\phi(\W^{(1)}\x)),\quad \W^{(i)} 
\in 
M(d_{i},d_{i-1}), 
\end{align}
where $\phi$ is some activation function acting 
coordinate-wise. Letting 
$\cW=(\W^{(1)},\dots, \W^{(N)})$ and $\cV=(\V^{(1)},\dots, \V^{(N)})$, the 
corresponding squared loss function is  (with a slight abuse of notation) 
\begin{align}\label{def:jloss_multi}
\ploss(\cW) = \jloss(\cW,\cV ) = \bE_{\x \sim \cD}\left[ \frac{1}{2}\left( 
f(\x;\cW) - 
f(\x;\cV)  
\right)^2 \right].
\end{align}
It can be easily verified that for any $\cD$ and any $i = 2\dots N$, 
$\jloss$ remains invariant under pairwise transformations of the form 
$(\W^{(i+1)},\W^{(i)})\mapsto 
(\W^{(i+1)}(\P_{\pi})^\top, \P_{\pi}\W^{(i)})$ and 
$(\V^{(i+1)},\V^{(i)})\mapsto (\V^{(i+1)}(\P_{\pi})^\top, 
\P_{\pi}\V^{(i)})$ 
where $\pi\in S_{d_{i}}$ and $i\in[N]$. If, in addition, 
$\cV=(\bones^\top, \I_d,\dots,\I_d)$ and $\cD$ is invariant under 
permutations, then applying 
arguments similar to what is used in (\ref{sec:gamma_inv}), 
we see that 
$\ploss$ is also invariant under 
\begin{align}\label{eqn:multilayers_inv}
	(\W^{(2)},\W^{(1)})\mapsto (\W^{(2)}\P_\pi^\top,\P_\pi
	\W^{(1)} \P_\rho)
\end{align}
for any $\pi\in S_{d_1},\rho\in S_{d_{0}}$. An example for a critical 
point empirically found for this setting is provided in \pref{fig:6layers} 
below.
\todoc{Give the multi layers model + The invariance properties are tied 
with the compositional 
	structure ( can recover architecture 	}

\section{Empirical Results}\label{sec:empirical_res}
In this section, we aim to examine the isotropy of critical points detected
by plain SGD for various shallow ReLU networks. We note that providing a
comprehensive study of cases where the principle of least (or small) 
symmetry breaking holds is outside our scope. Rather, we report the 
isotropy type of \emph{approximate} critical points found empirically 
through the following procedure: In each experiment, we run 100 
instantiations of 
SGD with Xavier standard initialization until no significant improvement in 
the gradient norm is observed. \emph{In all experiments}, each SGD step is 
performed using a batch of $1000$ fresh	randomly generated samples and a 
fixed step size of $0.01$. 
Note that unlike the basic variant (\ref{def:ploss}) where some critical 
points are provably shown to be bad local minima (see a computer-aided 
proof in
\cite{safran2017spurious}), here we do not examine extremality 
properties. Lastly, we 
remark in passing that since gradients of invariant function are tangential 
to fixed subspaces (\pref{prop: grad}), one can find critical points of a 
given isotropy type by a suitable initialization. We shall not follow this 
approach here. \newcommand{\expd}{20}
\\

\begin{figure}[ht]
\begin{center}
	\includegraphics[scale=0.5]{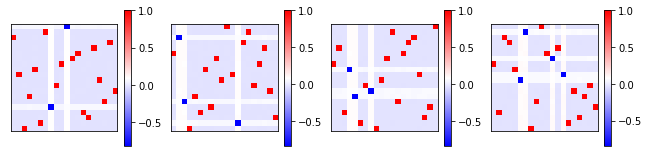}	
		\\
	\includegraphics[scale=0.5]{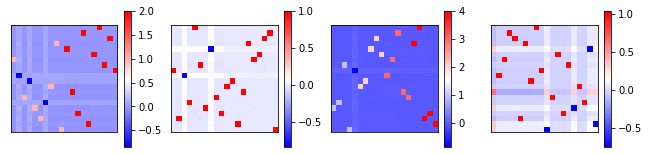}	
		
\end{center}
\caption{First row: approximate critical points of the two-layers ReLU 
net 
	(\ref{def:jloss}) with ground 
	truth $\V=\I_{20}$ and input distribution 
	$\cD=\cN(0,\I_{20})$. The 
	isotropy type (defined up to rows and columns 
	permutations) coincides 
	with the maximal isotropy groups of $\Delta S_{20}$ 
	(see, e.g.,
	\pref{fig:max_sym}). Second row: approximate critical points with 
	$(\V,\cD)$ set to 
	$(\I_{d/2} \oplus 2 	\I_{d/2}, \cN(0,\I_{20})), (\I_{20}, 
	\cN(0,\I_{d/2}\oplus \I_{d/2})), (\I_{d/4}\oplus 	
	2\I_{d/4}\oplus 3\I_{d/4}\oplus 4\I_{d/4}, 
	\cN(0,\I_{20}))$ and $(\I_{20},\cN(0,\I_{d/4}\oplus 2\I_{d/4}\oplus 
	3\I_{d/4}\oplus 
	4\I_{d/4}))$,
	presented from left to right. The observed isotropy 
	types 	adapt to the reduced joint invariance 
	of the target model and the underlying 
	distribution.}\label{fig:3basics}
\end{figure}

We start by considering the basic ReLU variant $\jloss$ (\pref{def:jloss})
with the identity target matrix. Following the same procedure 
described
above for $d=\expd$, it is seen in \pref{fig:3basics} that all 
approximate critical points match the first category of (`pixelated 
Mondrian-like') maximal isotropy subgroups addressed in 	
\pref{lem:max_subgropus_sn}. Next, we run the same experiment with target 
weight matrix $\V_2 = \I_{10} \oplus 2 \I_{10}$ and $\V_3= \I_{5}\oplus 
2\I_{5}\oplus 3\I_{5}\oplus 4\I_{5}$. The more elaborated structure of 
critical points observed in this case match large isotropy types of 
the different weight matrix. A similar, though not identical, phenomenon is 
observed when the target weight matrix is set to identity and the 
underlying input distribution is set to zero-mean Gaussian with correlation 
matrix 
$\V_1$ and $\V_2$ (see \pref{fig:3basics}). When setting $k=21,22$ and 
$\V=\I_{20}$, our findings match the empirical results reported by 
\cite{safran2017spurious}. Here again, the structure of the local minima 
is consistent with large isotropy subgroups of 	$S_{21}\times
S_{20}$. Lastly, to test the robustness of our hypothesis to smoothness and
flatness of the activation function, we replace the ReLU activation function
by Leaky-ReLU with parameter $0.01$ and Softplus, $z \mapsto 	
\frac{1}{\beta}\log(1+e^{\beta z})$, with $\beta=1$; This produces the same
isotropy types observed for $\jloss$ with $\V=\I_{20}$ (though convergence
rates seem to be typically slower).

\begin{figure}[ht]
			\begin{center}
		\includegraphics[scale=0.5]{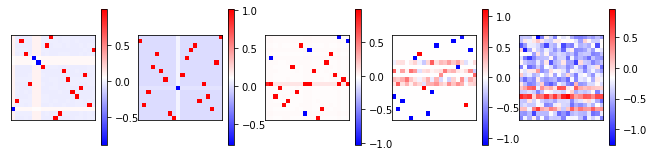}
	\end{center}
	\caption{Critical points of the ReLU variant defined 
	in 
		(\ref{def:ploss}) with an underlying distribution 
		$\cU([-1+C,1+C]^{20})$ where $C = 0,0.25,0.5,0.75,1$, 
		from left to 
		right. The observed isotropy types get 
		smaller when $C$ approaches~1.}
	\label{fig:4uniform}
\end{figure}

The least symmetry breaking principle \emph{does not} always 
hold. A concrete 
counterexample, for a setting where the observed isotropy 
types do 
not match the $\Delta S_d$-invariance, follows by letting the 
underlying 	
distribution to be $\cU([-1+C,1+C]^{20}), ~ 
C\in[0,1]$. As shown by \pref{fig:4uniform}, the symmetry of the critical 
points detected by SGD vanishes when $C$ approaches 1.
In other cases, the principle seems to hold only partially. 
For example, 
conducting the same experiment for
ReLU networks with a few fully-connected layers of constant width $d=20$ 
and target weights 
$\cV=(\bones_{20}^\top, \I_{20},\dots,\I_{20})$ (see 
\pref{def:jloss_multi}), shows that 
a similar structure is apparent only at the lower levels 
of the network, see \pref{fig:6layers}. 
\begin{figure}[ht]
\begin{center}
	\includegraphics[scale=0.4]{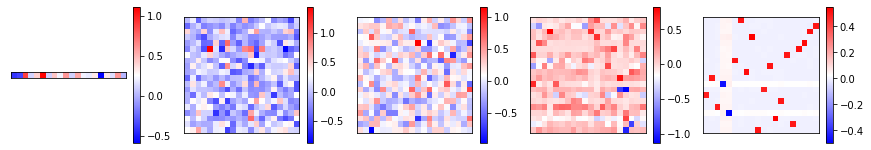}
	\includegraphics[scale=0.45]{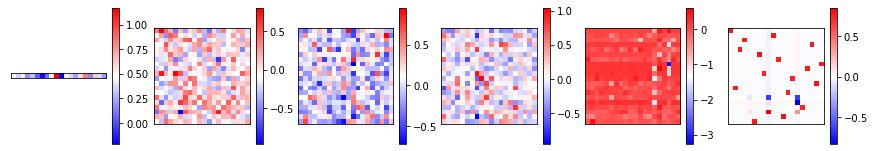}		
\end{center}
\caption{The weight matrices of four- and five- and six-layers 
fully-connected ReLU networks (the rightmost matrix corresponds to the 
bottom layer). The isotropy of the first and the second layers seem 
	to be consistent with the invariance properties shown in 
	(\ref{eqn:multilayers_inv}).}
\end{figure}\label{fig:6layers}

%

%

\section{Conclusion}
In this work, we demonstrate a definite effect of the joint symmetry
of various data distributions and shallow neural network
architectures on the structure of critical points. In particular, the
structure of spurious local minima detected by SGD in
\cite{safran2017spurious} are shown to be in a direct correspondence with
large isotropy subgroups of the target model, suggesting a new set of 
\emph{complementary} tools for studying loss landscapes associated with
neural networks. On the flip side, so far, we have not been able to provide 
a rigorous proof as to why local minima tend to be highly symmetric in the
settings considered in the paper. In addition, and perhaps more importantly,
it is not clear at the moment how wide the phenomenon of 
least symmetry breaking is (e.g., \pref{fig:4uniform} and 
\pref{fig:6layers}), or \emph{how well it relates to realistic settings} 
of, say, convents trained over image databases. \\

To conclude, the fact that a purely theoretical analysis gives a rather 
precise description of what is observed in practice, together with the fact 
that such phenomena have been encountered many times in various 
scientific fields are encouraging and form our main motivation for reporting
these findings. Of course, isotropy analysis 
only characterize the landscape of generic invariant 
functions to a limited degree (as further demonstrated by 
\cite{scheurle2015minima}); It is 
for this reason that we believe that the hidden mechanism, 
which makes this analysis suitable for the settings considered here, merits 
further investigation. 	\todoc{rethinking generalization,implicitly bias }

\section{Acknowledgments}
Part of this work was completed while YA was visiting the Simons Institute 
for the Foundations of Deep Learning program. We thank Haggai Maron, 
Michelle O'Connell, Nimrod Segol, Ohad Shamir, Michal Shavit, Daniel 
Soudry and Alex Wein for helpful and insightful discussions.

\bibliography{refs}
\bibliographystyle{plain}

\newpage
\appendix 

%
%
%
%

	\section{Proofs}
	\subsection{Proof for \pref{prop: grad}}
	\proof Let $Df: E \arr L(E,\real); \x \mapsto 
	Df_\x$ denote the derivative map of $f$. It holds that for all $\x,\e 
	\in E, g \in G$
	\begin{align}
	Df_\x(\e) 
	&= \lim_{t\arr 0} \frac{f(\x + t\e) - f(\e)}{t}\\
	&= \lim_{t\arr 0} \frac{f(g(\x + t\e)) - f(g\e)}{t}\\
	&= \lim_{t\arr 0} \frac{f(g\x + tg\e) - f(g\e)}{t}\\
	&= Df_{g\x}(g\e).
	\end{align}
	By definition of the gradient vector field we have for any $\e \in 
	\real^n$,
	\begin{align}
	\inner{\grad{f}(g\x),\e} 
	= Df_{g\x}(\e) 
	= Df_\x(g^{-1}\e) 
	= \inner{\grad{f}(\x),g^{-1}\e}
	= \inner{g^{-\top}\grad{f}(\x),\e}.
	\end{align}
	Hence for all $\e \in \real^n$, $\langle \grad{f}(g\x),\e\rangle = 
	\langle 
	g^{-\top}\,\grad{f}(\x),\e\rangle$ and so $\grad{f}(g\x) = 
	g^{-\top}\,\grad{f}(\x)$. In particular, if $G\subseteq \On{n}$, then 
	$g=g^{-\top}$ and $\grad{f}$ is $G$-equivariant.  The rest of the 
	properties are immediate corollaries of the $G$-equivariance of the 
	gradient.
	\qed

	\subsection{Full Derivation of \pref{eqn:glninv}} \label{sec:eqn_glninv}
	For notational convenience, we let $\U= \W^{(i)}$ and $\V= \W^{(i+1)}$ and 
	focus only on these two variables. Concretely, assume 
	$f(\A\U,\V\A^{-1})=f(\U,\V)$, for any $\A\in \GLn{\RR^n}$. Define $\Pi$ to 
	be the corresponding representation of $\GLn{\RR^n}$. Then,
	\begin{align*}
	\frac{\partial \Pi(\A)(\U,\V)}{\partial \A} =  \frac{\partial 
	(\A\U,\V\A^{-1})}{\partial \A} = 
	\mymat{\U^\top \otimes \I_{d_{i}} \\ - \A^{-\top}\otimes \V\A^{-1}}
	\end{align*}
	Thus,  
	\begin{align*}
	\left.\frac{\partial \Pi(\A)(\U,\V)}{\partial \A}\right|_{\A=\I}= 
	\mymat{\U^\top \otimes \I_{d_{i}} \\ - \I_{d_i}\otimes \V}.
	\end{align*}
	On the other hand
	\begin{align*}
	D \Phi(\U,\V ) = D\prn*{(\U\U^\top - \V^\top\V )/2} =	\mymat{\U^\top 
	\otimes \I_{d_{i}} \\ - \I_{d_i}\otimes \V}  (\I_{d_i^2}+\K_{d_i,d_i}),
	\end{align*}
	where $\K_{d_i,d_i}$ is the respective commutation matrix (see 
	\cite[Chapter 3.7]{magnus2019matrix}). The rest of the proof follows 
	mutatis mutandis.

	\subsection{Critical Points of \texorpdfstring{$\fofh$}{Eta}} 
	\label{sec:h_analysis}
	\newcommand{\nzset}{I}
	We consider the more general real-valued function $f:\RR^n\to\RR$  defined 
	as follows (see \cite[Section 4.5.3]{Field2007} for a more detailed study 
	of $\grad{\fofh}$ and $\HN$),
	\begin{align}
	f(\x) =  \frac{a}{2} \nrm{\x}^2 + \frac{b}{4}\nrm{\x}^4 + 
	\frac{c}{4}\sum^n_{i=1} {x_i^4}.
	\end{align}
	We have
	\begin{align}
	\grad{f}(\x) &= a\x + b\nrm{\x}^2\x + c \prn{x_1^3, \dots , 
	x_n^3}^\top,\\
	\hes{f}(\x) &= aI + b\|\x\|^2I + 2b \x\x^\top 
	+3c\Diag{x_1^2,\dots,x_n^2}. 
	\label{eqn:hes_f_pol}
	\end{align}
	Given $I\subset [n]$, we form a set of solutions whose 
	zero-coordinates are 
	specified by $\nzset$. Indeed, assuming $\frac{-a}{b|\nzset|+c}>0$, we 
	set 
	\begin{align*}
	x_i = \begin{cases}
	0 & i\in \nzset,\\
	\sqrt{\frac{-a}{b|\nzset|+c }}
	& i\notin \nzset.
	\end{cases}
	\end{align*}
	In particular, $\|\x\|^2=\frac{-a|\nzset|}{b|\nzset|+c}$. Therefore, 
	for 
	any 
	$i\in [n]$, 
	it follows that 
	\begin{align*}
	\grad{f}_i(\x) &= x_i(a + b\nrm{\x}^2 + c {x_i^2}) = x_i\prn*{ a - 
		\frac{ab|\nzset|}{b|\nzset|+c}  + c {x_i^2}}  = 
		\frac{x_i}{c}\prn*{\frac{a 
		}{b|\nzset|+c} 
		+  {x_i^2}}=0.
	\end{align*}
	Clearly, any change of sign of $x_i$ for some $i\in \nzset$ produces 
	a different critical point. It is straightforward to verify that these are 
	the only critical points which correspond to $S$. The overall number of 
	critical points is therefore $\sum_{i=0}^n \binom{n}{i}2^{n-i}=3^n$. 
	Moreover, it is 
	easy 
	to 
	verify that the isotropy group of any critical point 
	$\x\in\crl*{-\sqrt{{-a 	
			}/{(b|\nzset|+c)}} ,0,\sqrt{{-a }/{(b|\nzset|+c)}}}^n$ is 
			conjugate 
			to 
	$S_{p}\times H_{n-p}$, where $p$ is the number of non-zero coordinates 
	of 
	$\x$. 
	\\
	\\
	Next, we compute the eigenvalues of the Hessians at the critical 
	points of $f$  in order to determine their extremal properties. Let 
	$\x\in\crl*{-\sqrt{{-a 	}/{(b|S|+c)}} ,0,\sqrt{{-a }/{(b|S|+c)}}}^n$, with 
	at 	most $|S|$ nonzero entries, be a critical point and assume 
	w.l.o.g. that the nonzero coordinates appear before the zero 
	coordinates. Letting $M \in \RR^{|S|\times|S|}$ be defined by $M_{ij} = 
	\sgn(x_ix_j)$, 
	we 
	have by \pref{eqn:hes_f_pol} that 
	\begin{align*}
	\hes{f}(\x) &= aI + b\|\x\|^2I + 2b \x\x^\top 
	+3c\Diag{x_1^2,\dots,x_n^2}\\
	&= \prn*{a - \frac{ab|S| }{b|S|+c}}I - \frac{2ab }{b|S|+c} M\oplus 
	0_{n-|S|} 
	-\frac{3ac 
	}{b|S|+c}I_{|S|}\oplus 0_{n-|S|}\\
	&= \prn*{ \frac{ac }{b|S|+c}}I_{|S|}\oplus I_{n-|S|} - \frac{2ab 
	}{b|S|+c} 
	M\oplus 0_{n-|S|} 
	-\frac{3ac }{b|S|+c}I_{|S|}\oplus 0_{n-|S|}\\
	&= \frac{-a}{{b|S|+c}}
	\brk*{\prn*{ 2c  I_{|S|} + 2b M  } 
		\oplus -c I_{n-|S|} }.
	\end{align*}
	The eigenvalues of $M$ can be easily verified to be 
	$\{|S|,0,\dots,0\}$, 
	which implies that the eigenvalues of the Hessian are 
	\begin{align}
	\frac{-a}{{b|S|+c}}\crl*{2(b|S|+c),\underbrace{2c,\dots,2c}_{|S|-1\text{
				times}},\underbrace{-c,\dots,-c}_{n-|S|} }.
	\end{align}
	It follows then that if $b,c> 0>a$ then a critical point is extremal 
	iff its isotropy groups is $S_n$ or $H_n$, which corresponds to 
	a minima at $\sqrt{{-a}/{({b|S|+c})}}(\pm1,\dots,\pm1)$ or a maxima 
	at 	$\bzeros$, respectively. Additional examples for 
	critical points of other classes of finite reflections 
	groups, e.g., the symmetric group $S_d$, can be found 
	\cite{Field2007}.

	\section{Complementary for \pref{sec:main_res}}

	\subsection{Proof of \pref{lem:rect}} 
		$H$ acts transitively on the set of rows of $R_{a b}$. It follows that 
		if $R$ is a row of $R_{ab}$ and $h\in H$, then for all $\ell \in 
		[t_{ab}]$, $h:R\cap T_\ell^{ab}\to hR\cap T_\ell^{ab}$, and is a 
		bijection. A similar proof holds for columns.

	\subsection{Third Examples in 
	(\ref{ex:prod_not_prod})}\label{sec:not_diag_app}
	Recall our convention that the group $H_1$ permutes rows and $H_2$
	permutes columns. We write elements $(\pi,\rho)\in H_1 \times H_2$
	as $(\pi^r,\rho^c)$ to emphasize this and let $e^r$, $e^c$ denote the 
	identity elements of $H_1$, $H_2$, respectively. With these conventions, 
	observe that $\Gamma_\W$ contains the seven non-trivial symmetries
	\[
	\begin{matrix}
	((13)^c(24)^c,(12)^r) & ((12)^c(34)^c,(34)^r) & ((23)^c(14)^c, 
	(12)^r(34)^r)\\
	((23)^c,(13)^r(24)^r) & ((14)^c,(14)^4(23)^r) &  \\
	((1342)^c,(1324)^r) & ((3124)^c,(3142)^r) &
	\end{matrix}
	\]
	which generate a group $H$ of order $8$, which is not isomorphic to a 
	product (as in Lemma~\ref{lem:rect}). The action of $H$ on the set of 
	entries $w_{ij} = \alpha$ and $w_{ij} = \beta$ is transitive.
	If $\Gamma_\W(1,1)$ contains component $w_{ij}$ with a $\beta$-entry, 
	then $\Gamma_\W$ must act transitively on $\is{4}^2$, which violates 
	our assumption that $\alpha \ne \beta$.  
	Therefore the only way the order of $\Gamma_\W$ can be greater than 
	$8$ is if there exists $h \in \Gamma_\W$ which fixes $(1,1)$ but is 
	not the identity. Necessarily, such an $h$ 
	fixes column 1 and row 1 and preserves $\alpha,\beta$ in the 
	complementary $3 \times 3$-matrix. However, this matrix is easily seen 
	to have trivial isotropy
	(within $M(3,3)$) and so $\Gamma_\W = H$. 
	One can verify directly that $\Gamma_\W$ is isomorphic to $\mc{D}_4$. 
	
	\subsection{Proof of \pref{lem:equal_diagonal}}

	\proof The first statement is immediate since $H$ is transitive and so 
	for 
	$i \in \is{k}$,  there exists $\sigma \in H$ such that $\sigma(1) = 
	i$. 
	Hence
	$(\sigma,\sigma)(1,1) = (i,i)$.  The converse follows since the 
	hypotheses imply that each row and column contain exactly
	one point in the $\Gamma_\W$-orbit of $(i_0,j_0)$. Hence we can 
	permute rows and columns so	that the diagonal entries are 
	identical and 
	differ from the off-diagonal entries (use 
	Lemma~\ref{lem:rect}). It is now 
	easy to	see that the isotropy of the permuted matrix is of 
	diagonal 
	type---the conjugacy with $\Gamma_\W$ is given by the 
	permutation	that makes the diagonal entries equal. 
	\qed
	
	\subsection{Proof of \pref{lem:max_subgropus_sn}} 
	\proof (Sketch) (1) If $p = q = d/2$, we can add all 
	permutations 
	which map 
	$\is{p}$ to $\is{d} \smallsetminus \is{p}$ to obtain a 
	larger proper 
	transitive subgroup of $S_d$. 
	(2) The idea here is that the transitive partition breaks into $q$ 
	blocks 
	$(B_i)_{i\in \is{q}}$ each of size $p$. Elements of the wreath product 
	act 
	by permuting elements 
	in each block and then permuting the blocks (the basic wreath product 
	structure).  The order of the group is $|S_p|^q|S_q| = (p!)^q q!$.  We 
	refer the reader 	to~\cite{Liebecketal1987} for details and references.
	\qed


\end{document}